\documentclass[letterpaper, 10 pt, conference]{ieeeconf}  

\IEEEoverridecommandlockouts                              

\usepackage{graphics} 
\usepackage{epsfig} 
\usepackage{mathptmx} 
\usepackage{times} 
\usepackage{amsmath} 
\usepackage{amssymb}  
\usepackage{todonotes}
\usepackage{subcaption}
\usepackage{multirow}
\usepackage{gensymb}
\usepackage{ifthen}
\usepackage{bm}
\usepackage{cite}
\usepackage{siunitx}
\makeatletter
\let\NAT@parse\undefined
\makeatother
\usepackage{hyperref}
\DeclareMathAlphabet{\mathcal}{OMS}{cmsy}{m}{n}
\newcommand{\dataset}[1]{\texttt{#1}}

\title{\LARGE \bf
Towards introspective loop closure in 4D radar SLAM
}

\author{Maximilian Hilger, Vladimír Kubelka, Daniel Adolfsson, Henrik Andreasson, Achim J. Lilienthal$^{*}$
\thanks{This work was supported by the EU H2020 program (ITN project MORE, No. 858101) }
\thanks{All authors are with the RNP lab of the AASS research centre, \"Orebro, Sweden. Corresponding author:
        {\tt\small maximilian.hilger@oru.se}}%
\thanks{$^{*}$Achim J. Lilienthal's main affiliation is with the Chair Perception for Intelligent Systems at TU Munich}%
}

\begin{document}

\maketitle
\thispagestyle{empty}
\pagestyle{empty}

\begin{abstract}
Imaging radar is an emerging sensor modality in the context of Localization and Mapping (SLAM), especially suitable for vision-obstructed environments.
This article investigates the use of 4D imaging radars for SLAM and analyzes the challenges in robust loop closure.
Previous work indicates that 4D radars, together with inertial measurements, offer ample information for accurate odometry estimation.
However, the low field of view, limited resolution, and sparse and noisy measurements render loop closure a significantly more challenging problem.
Our work builds on the previous work - TBV SLAM – which was proposed for robust loop closure with 360$^\circ$ spinning radars. This article highlights and addresses challenges inherited from a directional 4D radar, such as sparsity, noise, and reduced field of view, and discusses why the common definition of a loop closure is unsuitable.
By combining multiple quality measures for accurate loop closure detection adapted to 4D radar data, significant results in trajectory estimation are achieved; the absolute trajectory error is as low as 0.46$\,$m over a distance of 1.8$\,$km, with consistent operation over multiple environments.
\end{abstract}

\section{INTRODUCTION}
Simultaneous Localization and Mapping (SLAM) is an essential part in today's autonomous mobile robots, enabling intelligent perception and navigation. Established sensing modalities, such as LiDAR and cameras, are failure-prone in vision-obstructed environments. In recent years, radar-based approaches have been proposed to overcome these challenges. However, traditional radar sensors measure only two-dimensional geometries (range and azimuth). With the introduction of 4D imaging radars, rich three-dimensional geometric information (range, azimuth, elevation) can be obtained with radar sensors for the first time. In addition, the \emph{4D} stands for measuring the relative velocity of each point through the analysis of the returning wave phase shift (further denoted as \emph{Doppler} measurements). Still, the point clouds are subject to more noise and sparseness compared to spinning radars. This makes tasks which require accurate geometric information, such as scan matching and loop closure detection, challenging.

\begin{figure}[t!]
	\centering
	\includegraphics[width = 0.45\textwidth, trim = 35cm 22cm 35cm 25cm, clip]{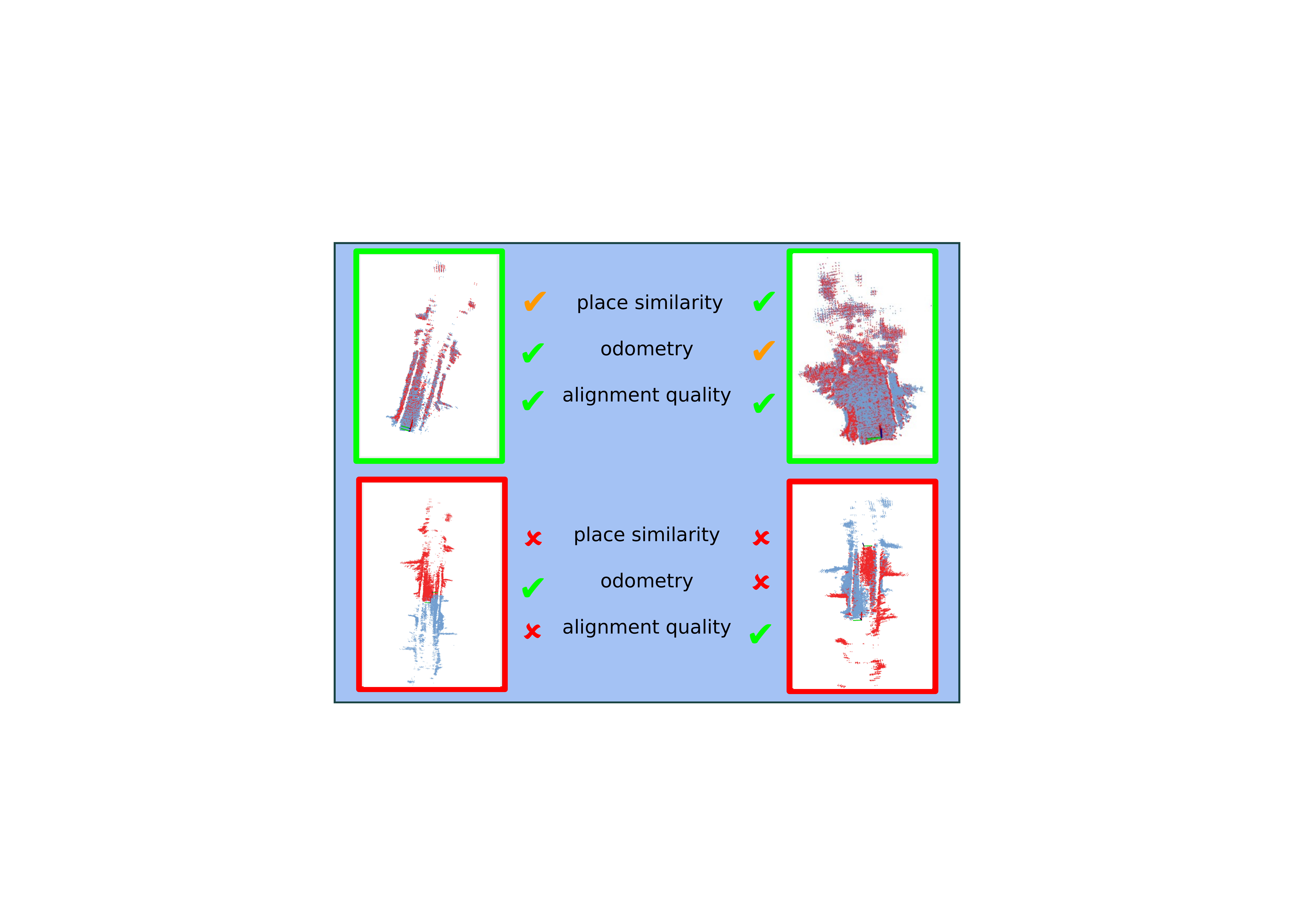}
	\caption{Top: Same-direction loop closures can be detected using multiple quality measures. Bottom: Opposite-direction loop closures cannot be detected.}
	\label{fig:examples}
    \vspace{-4mm}
\end{figure}

Initial work on 4D radar SLAM 
\cite{Zhang.2023, Li.2023, Zhuang.2023} applies the ScanContext descriptor \cite{kim.2018} for loop closure detection, which was originally developed for spinning LiDARs and radars. The addition of loop closures consistently improves the obtained pose estimation accuracy. However, no detailed analysis on the accuracy of the loop closure detection has been carried out so far. 

In this work, we focus on the accuracy that can be achieved with existing loop closure detection techniques. We investigate how different parameters influence the loop closure performance and discuss the occurring success and failure cases. For this, we use the state-of-the-art spinning radar SLAM framework TBV SLAM \cite{Adolfsson.2023}, which combines multiple quality measures for accurate loop closure detection, and adapt it to 4D radar data.
In our implementation, the source of the motion prior is a Doppler-Inertial odometry module \cite{Kubelka.2024} that exploits the Doppler measurement information.
The loop closure performance is evaluated on a radar dataset \cite{Kubelka.2024} recorded in underground and forest environments.
We show that for the same-direction loop closures, the combination of odometry similarity, \emph{ScanContext}~\cite{kim.2018}, and learnt alignment quality measures leads to significant improvement in trajectory estimation.
On the other hand, the experiments also show that detecting opposite-direction loop closures remains a challenge.

\section{RELATED WORK}
\label{sec:sota}
The rapid development of 4D radars has lead to the development of many radar SLAM systems and radar odometry methods which are an integral part of SLAM, providing an initial motion guess prior the loop-closure identification and optimization steps.

\subsection{4D radar odometry}
Existing 4D radar odometry methods can be loosely grouped into registration-free and registration-based methods. 
Registration-free approaches \cite{Doer.2020,Kramer.2020,Doer.2021,Ng.2021,Galeote.2023,Huang.2023} use the relative velocities measured by the 4D radar to estimate the ego-motion of the sensor. The position is then estimated by integrating the velocity over time, often in combination with an Inertial Measurement Unit (IMU). 
Registration-based odometry \cite{Retan.2021,Retan.2022,Michalczyk.2022,Michalczyk.2023,Lu.2024} uses the spatial information provided in the radar points to establish correspondences to previously recorded maps or scans. 
Recently, Kubelka et al. \cite{Kubelka.2024} compared multiple registration-free and registration-based approaches. They conclude that the choice of the radar odometry depends on the employed sensor configuration.
More specifically, highly accurate Doppler measurements supported by a low-drift IMU favors the registration-free approaches.
On the other hand, registration-based approaches are more suitable for radars with longer target persistence in subsequent radar scans and when using lower-grade IMUs.

\subsection{4D radar SLAM}
Building on top of the radar odometry systems, several SLAM approaches have been proposed. Zhang et al. \cite{Zhang.2023} employ scan-to-scan matching using adaptive probability distribution generalized ICP (APDGICP) that incorporate the uncertainty in the measured radar points. They detect loop closures using a combination of intensity ScanContext \cite{Wang.2020}, odometry, and barometric measurements. Li, Zhang, and Chen \cite{Li.2023} use scan-to-submap NDT matching to create radar odometry estimates. In a pose graph, they combine odometry factors with preintegrated ego-velocity factors and loop closure factors. The loop closure factors are obtained using ScanContext \cite{kim.2018}, followed by scan registration. Zhuang et al. \cite{Zhuang.2023} combine a 4D radar with an IMU in an iterative extended Kalman filter (IEKF) for odometry estimation. To detect loop closures, ScanContext \cite{kim.2018} and generalized ICP matching are used. In their follow-up work \cite{Wang.2023}, they accumulate multiple scans into place recognition descriptors. The authors report improved accuracy, although the effect was not quantified. 

\section{METHODOLOGY}

\begin{figure*}[t!]
	\centering
	\includegraphics[width = 0.9\textwidth]{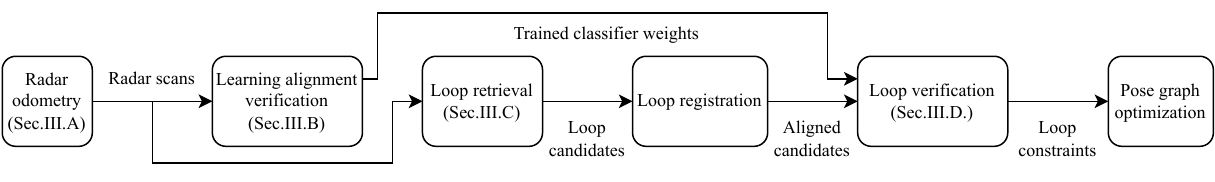}
	\caption{Architecture of the radar SLAM pipeline}
    \vspace{-5mm}
	\label{fig:pipeline}
\end{figure*}

Our framework extends TBV SLAM \cite{Adolfsson.2023b}, a state-of-the-art radar SLAM for spinning radar. The framework can be divided in four distinct components: 
\begin{enumerate}
    \item Odometry estimation
    \item Place recognition
    \item Registration and verification of loop candidates
    \item Pose graph optimization
\end{enumerate}
We specifically adjusted the place recognition and loop closure verification components to 4D radar point clouds. The used radar    odometry is described in section \ref{subsec:odom}. Using the odometry estimate, we learn to classify correctly aligned point clouds, detailed in section \ref{subsec:coral}. Possible loop candidates are then retrieved using a variation of ScanContext as explained in \ref{subsec:placerec}. Finally, in section~\ref{subsec:verification} we describe how the candidates are verified using multiple quality measures. An overview over the entire framework is given in Fig.~\ref{fig:pipeline}.  

\subsection{Odometry estimation}
\label{subsec:odom}
For odometry estimation, we use the "Doppler+IMU" odometry proposed in \cite{Kubelka.2024}. This method estimates the linear velocity of the robot using three-point RANSAC on the measured Doppler velocities of the radar points. The velocity is rotated into the world frame using the orientation provided by an IMU. The final pose is obtained by integration of the velocity. In the \dataset{mine-and-forest-radar-dataset}, this method is shown to outperform scan-matching based methods.

For the loop closure detection, keyframes are extracted from the trajectory every \SI{1.5}{\meter} or \SI{5}{\degree} as proposed in TBV SLAM \cite{Adolfsson.2023}. For each keyframe, we store the inlier pointcloud provided by RANSAC in the velocity estimation. Additionally, we compute three-dimensional oriented surface points, as introduced in CFEAR Radarodometry~\cite{Adolfsson.2023b}.

\subsection{Alignment verification}
\label{subsec:coral}
For assessing the point cloud alignment quality, we employ the framework CorAl (Correctly Aligned?) \cite{Adolfsson.2022}. 
The framework can be used to learn binary classification of alignment correctness between point clouds. 
To do that, complementary measures of alignment quality are computed and forwarded to a classifier. Training labels are automatically computed using odometry estimates.
In this work, we use measures of differential entropy and overlap, i.e. the measures $H_s$, $H_j$, and $H_o$ in the article ~\cite{Adolfsson.2022}.
While these entropy-based measures are suitable for quantifying small alignment errors, larger errors -- that are commonly present in loop closure detection -- are easier to capture with measures that are minimized in registration~\cite{Adolfsson.2023}. Hence, the entropy measures are supplemented with measures obtained from registration during loop closure. Specifically, we used the registration cost $C_f$, the average point cloud set size $C_a$, and the number of correspondences $C_o$. For a detailed description of these values, we refer the reader to \cite{Adolfsson.2022} and \cite{Adolfsson.2023}.

Following the procedure established in TBV SLAM~\cite{Adolfsson.2023}, we use the odometry prior to train the alignment classifier. Two consecutive radar scans are used to generate training samples. The transformation between the scans provided by the odometry is used to generate positive examples. Incorrect training examples are synthesized by introducing additional disturbances in the transformation before evaluating the alignment metrics. Small (\SI{0.5}{\meter}, \SI{0.5}{\degree}), medium (\SI{1}{\meter}, \SI{2}{\degree}), and large (\SI{2}{\meter}, \SI{15}{\degree}) disturbances are considered. Finally, a logistic regression classifier 
\begin{align}
p_{align} = \frac{1}{1+e^{-d_{align}}}, \: d_{align} = \bm{\beta} [H_j \: H_s \: H_o \: C_f \: C_o \: C_a \: 1]^T    
\end{align}
is used to discriminate between aligned and misaligned point clouds. 

\subsection{Place Recognition} 
\label{subsec:placerec}
For Place Recognition, we use the well-adopted ScanContext descriptor \cite{kim.2018}, which is based on a polar representation of the radar scan. First, we transform the inlier point cloud into a robot-fixed frame, which is parallel to the ground plane. After that, the environment of the robot is discretized to form an descriptor $\mathbf{I}_{n_{ring} x n_{secs}}$. Here, $n_{ring}$ expresses the number of range bins, and $n_{secs}$ is the number of azimuth bins. In contrast to \cite{Zhang.2023}, we consider the full 360° of the descriptor to prevent aliasing errors. Several encoding functions for ScanContext have been proposed: The original work uses the maximum height within a descriptor cell $z_{max}$ as value \cite{kim.2018}. In radar place recognition, the elevation can be subject to heavy noise. Hence, using the maximum intensity $p_{max}$ is preferred \cite{Zhang.2023}. However, the radar used in this work returns many ground points with intensity values similar to the reflections originating from other targets. Using the maximum intensity would discard the free-space information in front of the sensor. Thus, we adopt the encoding function proposed in \cite{Adolfsson.2023}: The values of the descriptor $\mathbf{I}$ are calculated as the sum of intensities of all points within a scan cell, divided by an occupancy-information-balancing weight (1000 in our experiments). Empty cells are set to a value of $\mathbf{I}(i,j)=-1$. Using this encoding function, vertical structures in the sensor's FoV can be leveraged as dominant features. To combat the sparsity and limited field of view of the radar point clouds, multiple consecutive keyframes can be used to increase the density of the descriptor. In this work, we do not consider origin augmentation, as the encountered loops are not laterally displaced.

In the loop retrieval step, we employ a coupled odometry/appearance matching approach \cite{Adolfsson.2023}. Given a query descriptor $\mathbf{I}^q$, we search for one or more loop candidates $c$ with descriptor $\mathbf{I}^c$ that minimize a joint retrieval cost
\begin{align}
c = \underset{c \in \mathcal {C}}{\operatorname{arg min}} \:d_{sc}(\mathbf {I}^{q},\mathbf {I}^{c}) + d^{q,c}_{odom},
\end{align}
where $d^{q,c}_{odom}$ penalizes large deviations from the odometry prior. 
This joint retrieval cost leverages the accurate odometry and improves the retrieval in ambiguous environments compared to descriptor-only retrieval. 

\subsection{Verification of loop candidates}
\label{subsec:verification}
After a loop candidate is retrieved, it is registered to the query scan using the oriented surface points and the p2d metric ~\cite{Adolfsson.2023b}. Following, a verification combining three different metrics is carried out:
\begin{enumerate}
    \item Odometry similarity $d_{odom}$
    \item ScanContext descriptor distance $d_{sc}$
    \item CorAl quality $d_{align}$
\end{enumerate}
Similar to the alignment verification, we use a logistic regression classifier to verify if a loop closure has been accepted:
\begin{align}
    &y_{loop}^{q,c_{k}} = \frac{1}{1+e^{-{\bm {\Theta} }\mathbf {X}^{q,c_{k}}_{loop}}}, \:\:\text{s.t.}\:\: y_{loop}^{q,c_{k}}> y_{th}, \\ &\mathbf {X}^{q,c_{k}}_{loop} = [d_{odom} \: d_{sc} \: d_{{ali}{gn}} \: 1]^{T}.
\end{align}
Multiple loop candidates can be verified to select the best possible constraint. The best scoring loop closure above the decision threshold $y_{th}$ is added to the pose graph.

\section{EVALUATION}

\begin{figure*}[t!]
	\centering
	\begin{subfigure}[b]{0.23\textwidth}
		\includegraphics[width = 1\textwidth]{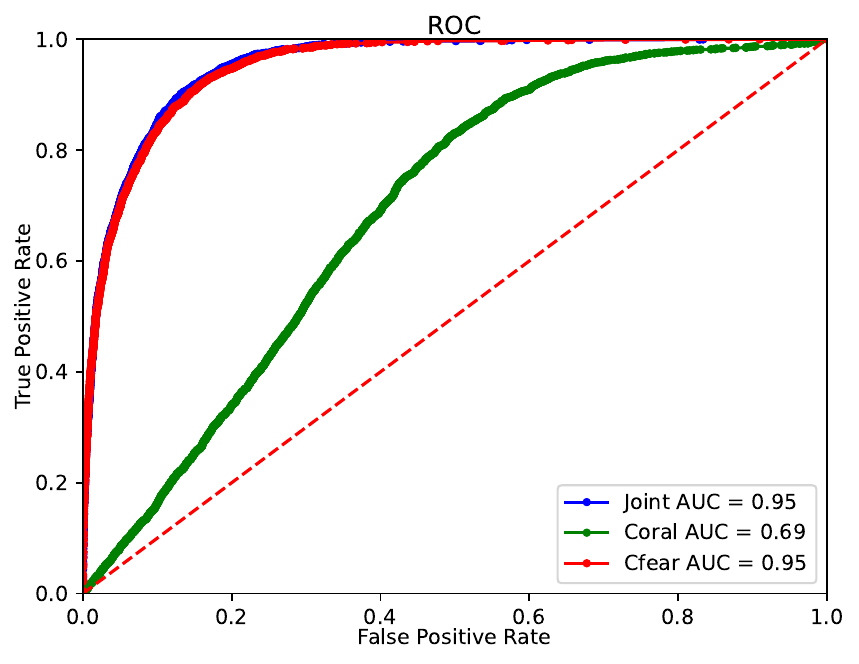}
		\caption{Scan alignment ROC curves, \dataset{Mine}.}
		\label{subfig:coralmine}
	\end{subfigure}
	\begin{subfigure}[b]{0.23\textwidth}
		\includegraphics[width = 1\textwidth]{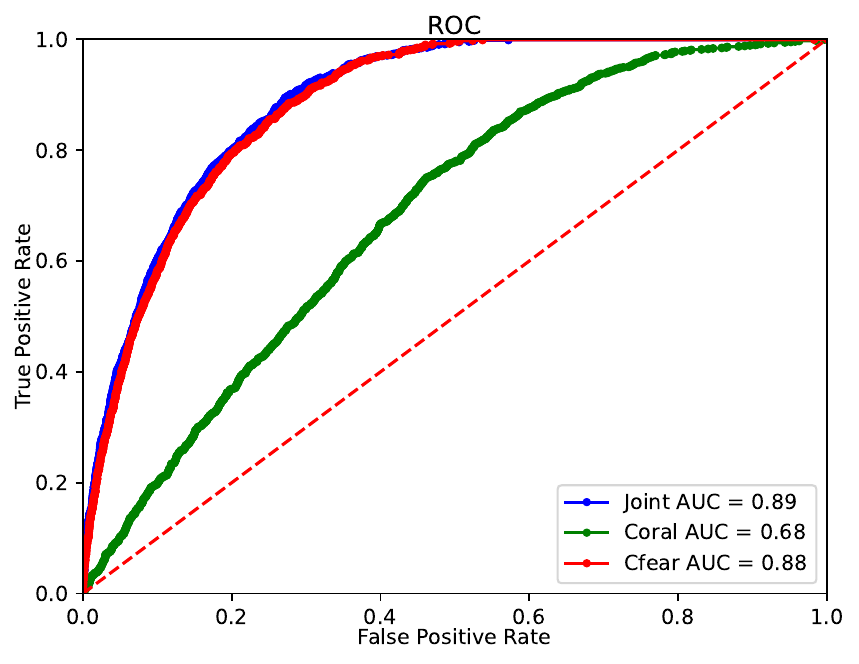}
		\caption{Scan alignment ROC curves, \dataset{Forest}.}
		\label{subfig:coralforest}
	\end{subfigure}
	\begin{subfigure}[b]{0.23\textwidth}
		\includegraphics[width = 1\textwidth]{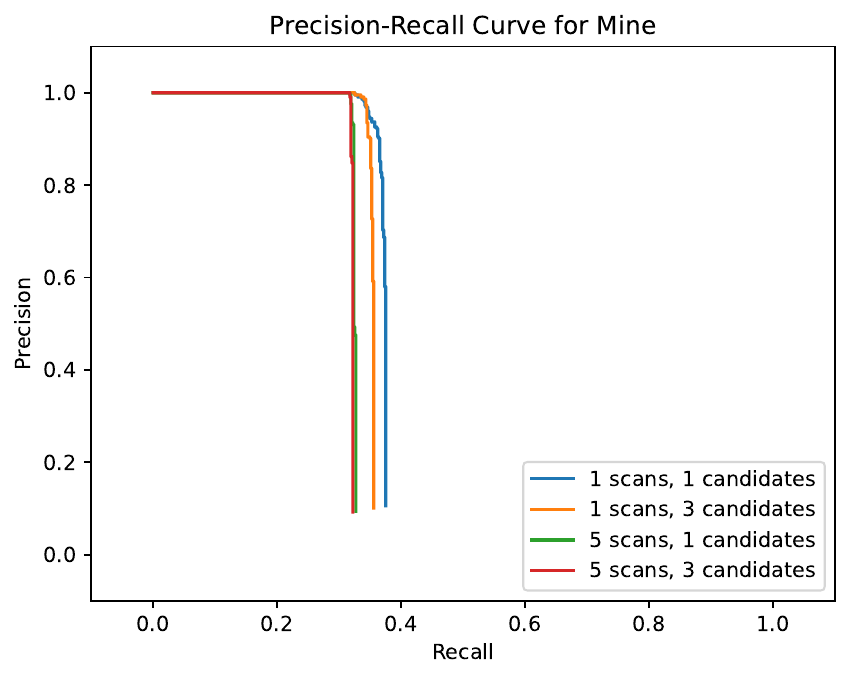}
		\caption{Loop verification PR curves, \dataset{Mine}.}
		\label{subfig:prmine}
	\end{subfigure}
	\begin{subfigure}[b]{0.23\textwidth}
		\includegraphics[width = 1\textwidth]{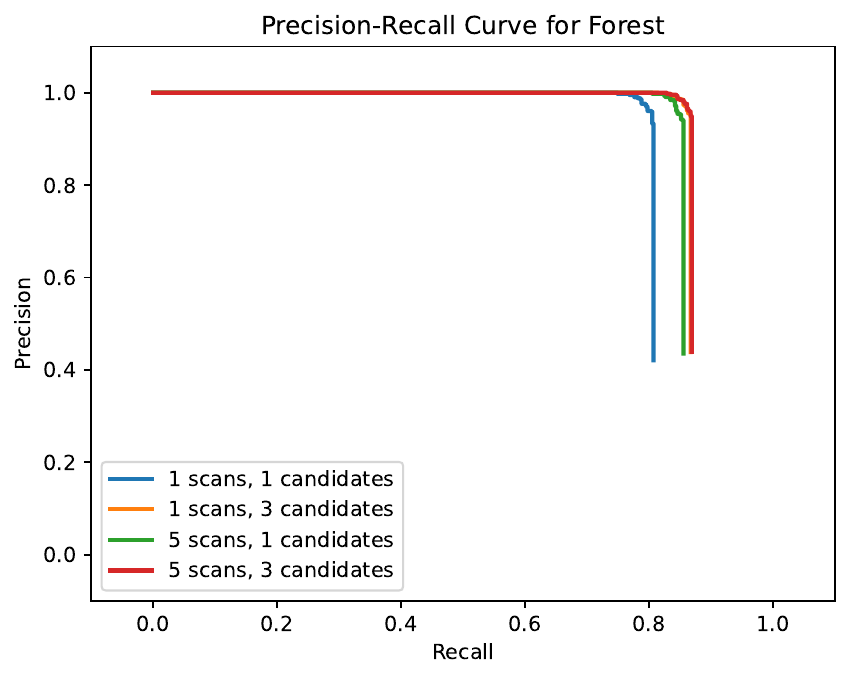}
		\caption{Loop verification PR curves, \dataset{Forest}.}
		\label{subfig:prforest}
	\end{subfigure}
	\caption{Classifier evaluation for alignment classification and loop closure evaluation}
	\label{fig:evaluation}
\end{figure*}

We proposed a framework for loop closure detection using 4D radar sensors. In this section, we evaluate our three main claims: (i) Misalignment can be classified in 4D radar scans, (ii) Loop closures can be detected and verified using multiple quality metrics, (iii) the detected loop closures lead to increased accuracy in the trajectory estimation.

\subsection{Evaluation metrics}
For evaluation of the loop closures accuracy, we need to label ground truth loop closures. A standard approach is to label all scans taken at a similar position as true loops. This approach, however, leads to non-detectable loops if the sensor readings do not overlap, as it can be the case for 4D radars. While this is not a problem as long as the loops are traversed in the same direction, revisits in opposite directions cannot be properly detected. However, to remain consistent with existing literature and to emphasise the need for more accurate metrics, we consider a loop closure as true as long as the two sensor poses are not farther away than $6~m$.

\subsection{Experimental setup}
We use the \dataset{mine-and-forest-radar-dataset} \cite{Kubelka.2024} to evaluate our approach. It consists of two sequences, \dataset{Mine}, and \dataset{Forest}. While the \dataset{Forest} sequence is traversing a large loop two times in the same direction, the \dataset{Mine} sequence is offers also loop traversals in reverse direction. The dataset is recorded using a Sensrad Hugin A3 radar and a Xsens MTi-30 IMU.

We split the evaluation in three parts: First, we evaluate the classification of the alignment quality. Here, we compare three different classifiers based on single scans: 1) CorAl measures 2) CFEAR p2d measures 3) CorAl measures + CFEAR p2d measures. We train one classifier for each environment.

In the second part, we evaluate the loop closure detection. For each sequence, we execute 3 experiments: 
\begin{enumerate}
    \item constructing the ScanContext descriptor out of only one keyframe 
    \item accumulate 5 keyframes into one descriptor
    \item verify top 3 candidates
    \item accumulate 5 keyframes and verify top 3 candidates.
\end{enumerate}
Lastly, we evaluate how the detected loop closures increase the quality of the estimated trajectories. We use the best scoring loop classifier from the second evaluation step.

\subsection{Performance evaluation}
\textit{1) Alignment classification:} To evaluate the performance of the alignment classification, we use the ROC curve and the AUROC score. These are given in Fig,~\ref{subfig:coralmine}-\ref{subfig:coralforest}, respectively.

We observe that the entropy-based CorAl quality measures do not work well inour 4D radar.
As elaborated in \cite{Adolfsson.2022}, CorAl's accuracy can be impaired if patches with low point density occur.
Due to numerical issues, even joining well-aligned point clouds may lead to a large increase in entropy, which is hard to distinguish be´from an increase caused by misaligned point clouds.
This problem was already observed with spinning 3D LiDAR, but appears to be even more prevalent in 4D radar point clouds. In all sequences, the false positive rate is significantly higher. In addition, the ROC curves of the CFEAR-only classifier and the combined classifier align, which means that the combined classifier learns to ignore the CorAl measures. Hence, we only consider the CFEAR-only quality metric for the remainder of this work.

\textit{2) Loop closure detection:}
For the evaluation of the loop closure detection, we use the precision-recall curves given in Fig.~\ref{subfig:prmine}-\ref{subfig:prforest}. Here, we analyse how different submap sizes for descriptor generation and multi-candidate retrieval influence the performance of the loop closure detection.

\begin{figure}
    \centering
    \includegraphics[width = 0.28\textwidth]{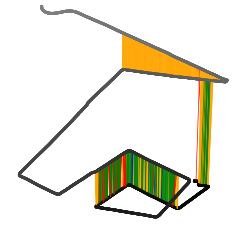}
    \caption{Detected loops in \dataset{Mine} sequence using 5 keyframes for descriptor generation and top 3 retrieved candidates. Red (dangerous failure – false loop with undesired high confidence), orange (safe failure – false loop but with desired low confidence), blue (safe failure, correct loop with undesired low confidence), green (success, correct loop with desired high confidence). The failures in the upper part of the trajectory originate from the traversal in reverse direction.}
    \label{fig:loopsMine53}
\end{figure}

In the \dataset{Mine} sequence, we observe that the best scoring classifier only uses one scan for descriptor generation and verifies just a single guess. To further enhance the understanding of this failure mode, we additionally provide an illustration of the correctly and the incorrectly detected loop closures in Fig.~\ref{fig:loopsMine53}. Most false positive loop closures are detected along sections in the mine without side tunnels. These loops are particularly hard to detect due to the environment's uniformity. The alignment classification can hardly distinguish between correct and incorrect lateral displacement without sufficient features. This problem is stressed by the short-range setting of the employed radar sensor. Loop closures in reverse direction cannot be detected, but appear as safe failures and do not lead to false positives. A correctly detected loop closure is depicted in fig.~\ref{subfig:loopcorrect}, and an example for a missed loop due to insufficient overlap is given in fig.~\ref{subfig:loopmissed}.

In the \dataset{Forest} sequence, the loops can be closed accurately at a high true-positive rate. We observe, that both fusing multiple keyframes and retrieving multiple loop candidates improves the loop closure performance. However, retrieving multiple candidates has a larger influence. When using three retrieved candidates, the different number of keyframes does not influence the accuracy of the classifier anymore. This sequence underlines the potential of the TBV SLAM framework.

\begin{figure}
    \centering
    
	\begin{subfigure}[b]{0.23\textwidth}
		\includegraphics[width = 1\textwidth]{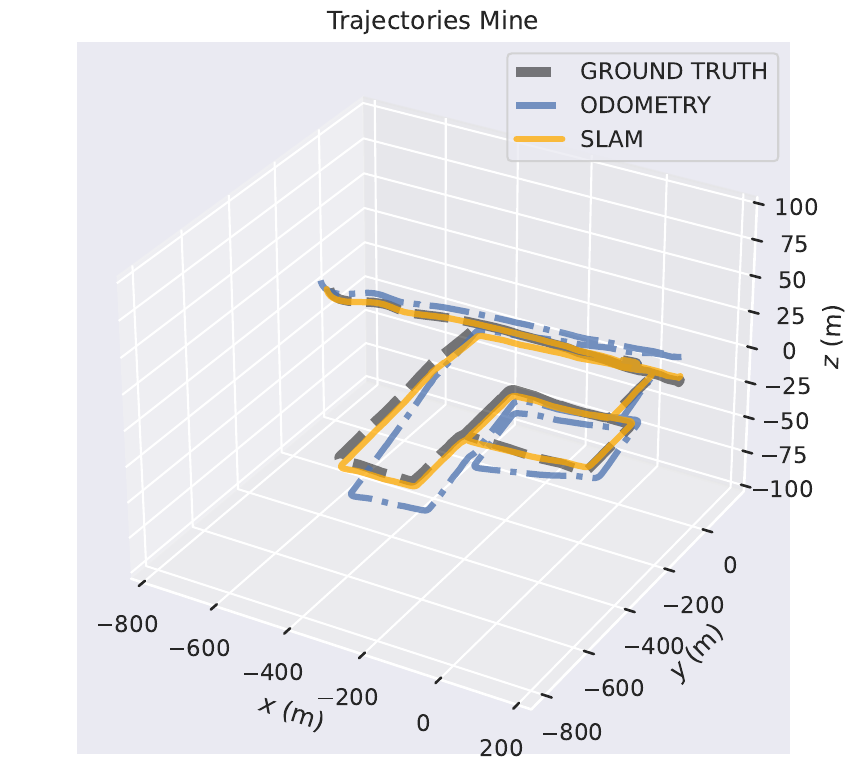}
		\caption{Trajectories, \dataset{Mine}.}
		\label{subfig:trajmine}
	\end{subfigure}
	\begin{subfigure}[b]{0.23\textwidth}
		\includegraphics[width = 1\textwidth]{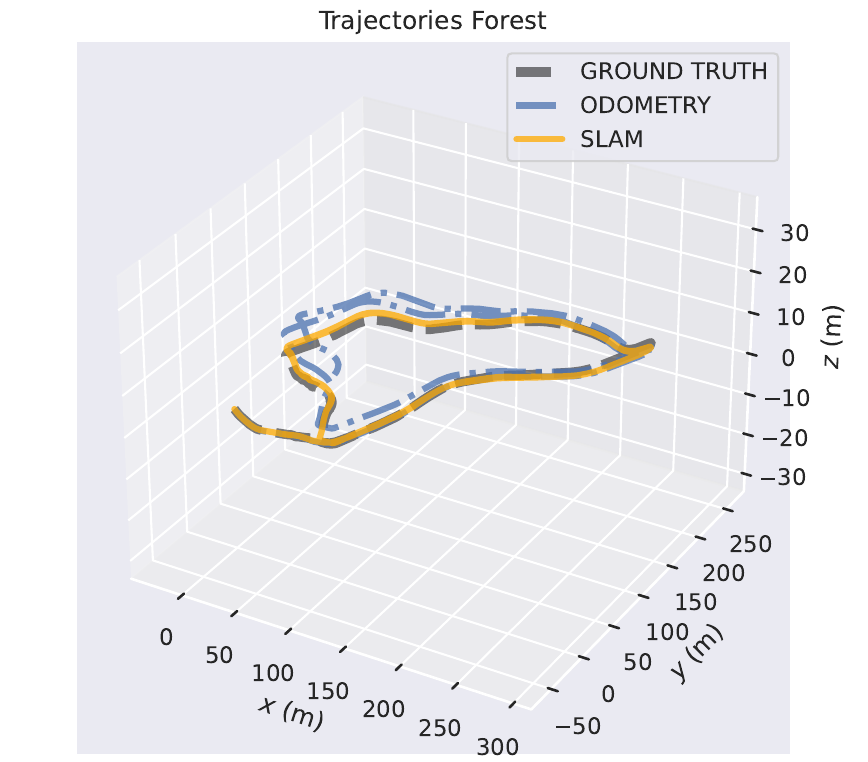}
		\caption{Trajectories, \dataset{Forest}.}
		\label{subfig:trajforest}
	\end{subfigure}
    \caption{SLAM and odometry trajectories with estimated loop closures. Z axis is zoomed in to better visualize z drift.}
    \label{fig:traj}
\end{figure}

\textit{3) Trajectory quality:}
To assess the quality of the estimated trajectories, we employ the KITTI odometry metrics \cite{Geiger.2013} and the Absolute Trajectory Error (ATE). While the Kitti odometry error quantifies the translational drift $t_{rel}$ and rotational drift $r_{rel}$ on a smaller scale, \textit{ATE} expresses the global consistency of the trajectory by computing the Root Mean Squared Error between predicted and true poses. We compare the closed-loop metrics against the baseline odometry.

The resulting error metrics are given in table.~\ref{tab:trajectorymetrics}. We observe, that the loop closure detection is able to reduce the drift by up to 64\%. In addition, the ATE can be reduced to \SI{0.46}{\meter} in the \dataset{Forest} sequence, which only includes same-direction loop closures. The resulting trajectories are plotted in Fig.~\ref{subfig:trajmine}-\ref{subfig:trajforest}.

All in all, we observe that the loop closure in 4D SLAM works sufficiently well to significantly improve trajectory quality. However, multiple open challenges remain: How to detect and evaluate loops in different directions and how to cope with feature-sparse environments.

\begin{table}[t!]
	\centering
	\caption{Trajectory performance metrics.}
	\begin{tabular}{c|c c c |c c c}
		\hline
		  ~  & \multicolumn{3}{c|}{Odometry} & \multicolumn{3}{c}{SLAM} \\
		  Metric  & $t_{rel}$ & $r_{rel} $ & \textit{ATE} & $t_{rel}$ & $r_{rel} $ & \textit{ATE} \\
		  Dataset  & [\%] & [\degree/100m] & [m] & [\%] & [\degree/100m] & [m] \\
		\hline
	    \dataset{Mine}    & 1.037 & 0.313 & 6.541 & 0.813 & 0.264 & 3.211 \\
        \dataset{Forest}  & 1.235 & 0.625 & 5.440 & 0.444 & 0.304 & 0.455 \\
		\hline
	\end{tabular}	
	\label{tab:trajectorymetrics}    
\end{table}

\subsection{Discussion}
\begin{figure}
    \centering
	\begin{subfigure}[b]{0.18\textwidth}
		\includegraphics[width = 1\textwidth]{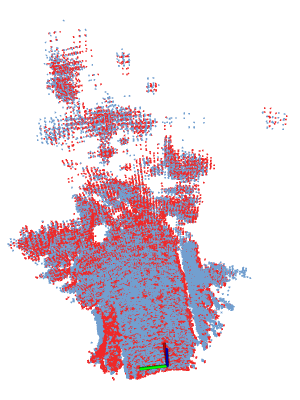}
		\caption{Correctly\\ detected loop closure.}
		\label{subfig:loopcorrect}
	\end{subfigure}
	\begin{subfigure}[b]{0.13\textwidth}
		\includegraphics[width = 1\textwidth]{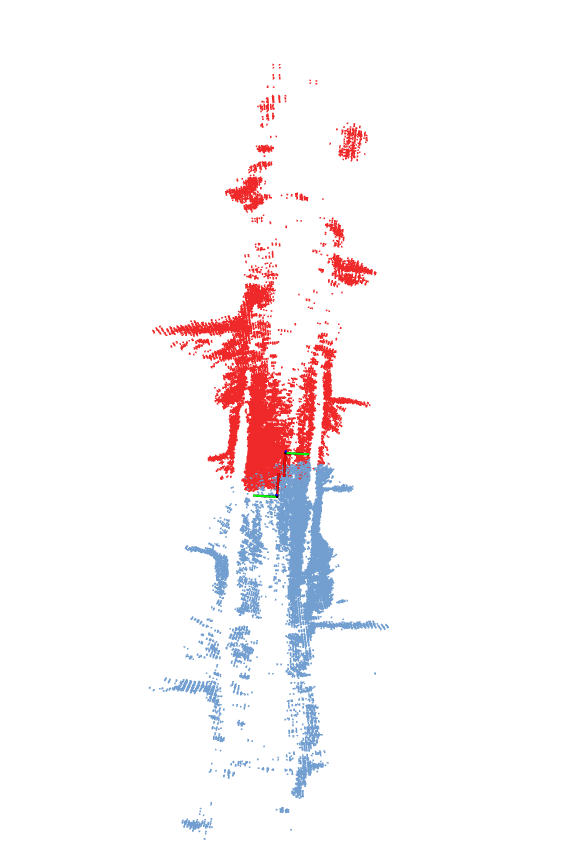}
		\caption{Missed loop, insufficient overlap.}
		\label{subfig:loopmissed}
	\end{subfigure}
	\begin{subfigure}[b]{0.13\textwidth}
		\includegraphics[width = 1\textwidth]{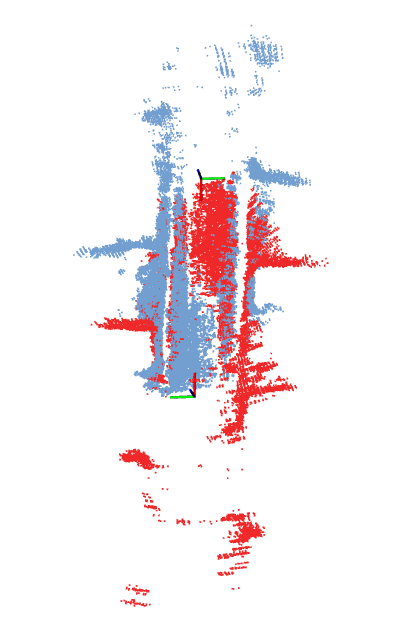}
		\caption{What should be considered as loop.}
		\label{subfig:loopdefinition}
	\end{subfigure}
    \caption{Examples for detected and missed loop closures. Blue: query scan, red: candidate scan.}
    \label{fig:loopPCLs}
\end{figure}

Finding loop closures in the opposite direction is the most striking problem found. To properly address this, it is not just necessary to improve the algorithms: More than that, the condition for two poses to be considered as participating in a loop closure need to be defined. Gupta et al. \cite{Gupta.2024} propose using the overlap of two scans as gate to consider a loop closure as true. An illustration for a possible loop closure given their definition is given in fig.~\ref{subfig:loopdefinition}. We believe that with this modified labeling approach and a modified loop retrieval component, highly-accurate loop closure detection and SLAM can be achieved with 4D radar.

\section{CONCLUSIONS AND FUTURE WORK}

In this paper, we analysed the common challenges in performing loop closure in 4D radar SLAM. We demonstrated that even though the TBV SLAM loop closure detection pipeline can be applied in 4D radar SLAM, only same-direction loop closures are detected correctly and lead to significant improvements in the trajectory estimation. To this end, we discussed how the common approaches to loop closure do not suit the use case of directional 4D radars. As for the future work, we aim to extend the experimental dataset and investigate how submap-based methods can be used to increase loop closure performance. Additionally, we want to assess how the trained classifiers generalize between environments and sensors.

\addtolength{\textheight}{-18cm}   


\bibliographystyle{IEEEtran}
\bibliography{IEEEabrv,References.bib}

\end{document}